\documentclass[sigconf]{acmart}

\AtBeginDocument{%
  }

\usepackage{multirow}

\setcopyright{none}
\renewcommand\footnotetextcopyrightpermission[1]{}
\begin{document}

\title{Practical Bayesian Inference for Speech SNNs: Uncertainty and Loss-Landscape Smoothing}

\author{Yesmine Abdennadher}

\affiliation{%
  \institution{Idiap Research Institute}
  \city{Martigny}
  \country{Switzerland}
}\email{yesmine.abdennadher@idiap.ch}

\author{Philip N. Garner}
\affiliation{%
  \institution{Idiap Research Institute}
  \city{Martigny}
  \country{Switzerland}}
\email{phil.garner@idiap.ch}

\renewcommand{\shortauthors}{Abdennadher and Garner}

\begin{abstract}
Spiking Neural Networks (SNNs) are naturally suited for speech processing tasks due to their specific dynamics, which allows them to handle temporal data. However, the threshold-based generation of spikes in SNNs intuitively causes an angular or irregular predictive landscape. We explore the effect of using the Bayesian learning approach for the weights on the irregular predictive landscape. For the surrogate-gradient SNNs, we also explore the application of the \textit{Improved Variational Online Newton} (IVON) approach, which is an efficient variational approach. The performance of the proposed approach is evaluated on the Heidelberg Digits and Speech Commands datasets. The hypothesis is that the Bayesian approach will result in a smoother and more regular predictive landscape, given the angular nature of the deterministic predictive landscape. The experimental evaluation of the proposed approach shows improved performance on the negative log-likelihood and Brier score. Furthermore, the proposed approach has resulted in a smoother and more regular predictive landscape compared to the deterministic approach, based on the one-dimensional slices of the weight space.
\end{abstract}


\maketitle
\section{Introduction}
Spiking Neural Networks are well suited for speech processing because they are fundamentally temporal in nature.  In principle, they can also achieve low-latency and energy-efficient ``event-driven'' inference.
Unlike traditional neural networks, in SNNs the communication between neurons is in the form of spikes, and the state is maintained using the dynamics of the neurons.

Nevertheless, training SNNs is not without difficulties. In practice, surrogate gradients enable the optimization of SNNs, where the derivative of the spike function is replaced by a smooth function during backpropagation. However, the forward pass is based on threshold-based spike generation, and this intuitively leads to an irregular optimization landscape. Specifically, this gives rise to the following question: Do deterministic SNNs result in a less regular predictive objective?

If this is true, then relying on a single deterministic estimate may not be optimal. In this case, a single estimate may not be informative about the weights and may not result in a highly robust predictive objective, especially when this is locally irregular. Therefore, it is important to look beyond deterministic training and consider distributions over weights.

In this work, we study \textit{Improved Variational Online Newton} (IVON) as a practical Bayesian learning method for surrogate-gradient SNNs. IVON maintains a Gaussian posterior approximation over the network weights and updates both the posterior mean and uncertainty during training. This is particularly appealing in the context of SNNs because learning under weight uncertainty can be interpreted as optimizing a locally averaged predictive objective, which intuitively smooths irregular behavior induced by threshold-based spike dynamics.

This leads to the central hypothesis of our work: if the local predictive landscape of a deterministic SNN is angular or piecewise irregular due to spike-threshold switching, then Bayesian learning over the weights may be especially beneficial. In particular, averaging predictions over a distribution of plausible weights may yield a smoother and less brittle predictive objective than deterministic point-estimate training.

We apply IVON on the Heidelberg Digits (HD) and Speech Commands (SC) benchmarks using predictive and uncertainty metrics, as well as one-dimensional weight-space slices. Our intention is to verify if the quality of the predictions is improved by the application of IVON and if the predictive landscape is smoother and more regular than in the case of deterministic training.

\paragraph{Contributions.}
We make three contributions: we examine IVON as an effective solution to weight uncertainty learning in surrogate gradient SNNs for speech recognition; we propose that IVON in SNNs is equivalent to regularizing the objective function of prediction by means of weight averaging in Bayesian methods; we provide supporting evidence in predictive, calibration, and local loss geometry analyses.

\section{Related work}
Bayesian methods aim to improve predictive reliability by modeling uncertainty over neural
network parameters. Many approaches include mean-field variational inference such as Bayes by
Backprop \cite{Graves2011,Blundell2015}, dropout-based approximations
\cite{GalGhahramani2016,Kingma2015}, stochastic-gradient MCMC methods \cite{WellingTeh2011,Chen2014},
and Gaussian approximations derived from curvature information or optimization trajectories
\cite{Maddox2019,Ritter2018,Daxberger2021}. Among post-hoc methods (i.e., applied after training), Laplace approximation is a
prominent approach \cite{Daxberger2021}. In practice, however, scalability often motivates fitting it
only on the last layer. In SNNs, this places uncertainty on the readout weights while keeping the
spiking feature extractor fixed, and therefore does not capture perturbations that modify the
internal spiking dynamics.

Variational methods are particularly attractive because
they maintain a weight distribution while remaining compatible with standard minibatch training \cite{khan2018,osawa2019,khan2023}.
IVON \cite{shen2024} belongs to this family and can be viewed as a Bayesian learning rule in which both parameter
means and uncertainty are updated jointly during training. It was introduced as a more
practical variant of earlier online variational Newton methods, with reported empirical improvements
over related approaches. 

In parallel, SNNs have been successfully applied to speech and audio tasks with
surrogate-gradient training \cite{BittarGarner2022,Wu2020DeepSNN_ASR}, but Bayesian learning in SNNs
remains relatively underexplored. Prior work has
mainly adapted variational Bayesian learning rules to spiking models
\cite{Skatchkovsky2022BayesianCL_SNN}. This question is also practically relevant beyond training:
in neuromorphic SNN deployments, effective parameter perturbations can arise from device mismatch,
fabrication variability, quantization, and other hardware non-idealities, and these effects have
been shown to degrade SNN inference and motivate robustness-oriented training
\cite{Buchel2021RobustDeployment,Cakal2023HardwareAware,Bhattacharjee2022CrossbarRobustness}. In
threshold-driven networks such as SNNs, even small parameter changes can alter spike timings or
spike occurrences and thereby affect predictions. Our focus is therefore narrower: we study IVON
specifically because its training-time weight perturbations are directly aligned with our central
question of whether learning under weight uncertainty can yield a smoother and less
perturbation-sensitive predictive objective.

\section{Task and data description}

We consider the problem of speech recognition, in which we wish to classify each short spoken utterance into one of a fixed number of classes. We are interested in two of the most commonly used benchmarks in previous work on this problem: \textit{Heidelberg Digits (HD)} and \textit{Speech Commands (SC)}. In both datasets, we begin with a raw audio waveform and compute a set of acoustic features, which are then input to the model as a sequence of feature vectors over a fixed window of time.

The HD dataset has spoken digits in English and German, with 20 classes, recorded from twelve different speakers, with two speakers only in the test set. The training set has 8,332 examples, the test set has 2,088 examples, and no validation set. The SC dataset uses Google Speech Commands v0.2, which has 35 keyword classes drawn from a diverse set of speakers. Using the standard split, the training set has 75,466 examples, the validation set has 9,981 examples, and the test set has 20,382 examples \cite{Warden2018}.

Although spiking versions of these datasets are also available (e.g., spike-based event representations specifically suited for training SNNs), we observed that in our pipeline, the feature-based (non-spiking) versions of these datasets produce stronger and more stable results for our speech recognition baselines \cite{BittarGarner2022}. We thus present our main results on HD and SC using the feature-based inputs and a unified evaluation protocol for both deterministic and Bayesian models.

\section{SNN Architecture and Training}

We use the speech input data to train the model, which is based on the feedforward spiking neural network (SNN) with leaky integrate and fire (LIF) neurons.

The structure of the network is composed of two hidden layers of LIF neurons and a spike-rate readout head. At any discrete point in time, denoted as $t \in \{1, \dots, T\}$, layer $\ell$ is provided with an input vector $x_t^{(\ell)}$ and applies a linear transformation followed by LIF membrane integration to produce a spike output. Each of the two hidden layers of LIF neurons is composed of a fully connected synaptic layer, batch normalization (BN) to ensure stable training, and dropout regularization with a given drop probability $p_{\mathrm{drop}}$. This is our proposed deterministic model, and it is also utilized as a backbone in our SNN models in conjunction with IVON in the Bayesian setting.

For the discrete-time LIF dynamics:
Let $u_t^{(\ell)}$ denote the membrane potential and $s_t^{(\ell)} \in \{0,1\}^{n_\ell}$ the spike vector at time $t$ in layer $\ell$. A standard discrete-time update is
\begin{align}
u_{t+1}^{(\ell)} &= \alpha^{(\ell)} u_t^{(\ell)} + W^{(\ell)} x_t^{(\ell)} - v_{\mathrm{th}}\, s_t^{(\ell)},\\
s_t^{(\ell)} &= \mathbb{I}\!\left[u_t^{(\ell)} \ge v_{\mathrm{th}}\right],
\end{align}
where $\alpha^{(\ell)} \in (0,1)$ controls the leak, $W^{(\ell)}$ are trainable synaptic weights, and $v_{\mathrm{th}}$ is a firing threshold. Since the spike function is non-differentiable, training relies on surrogate gradients, here using a boxcar surrogate function \cite{Neftci2019}, while keeping the forward dynamics strictly spiking.

For classification, we use a spike rate readout, which transforms the spike train of the last hidden layer into a single utterance-level prediction. We denote the spike vector of the last hidden layer at time step $t$ as $s_t^{(L)}$. To obtain the average firing rate over the fixed window of $T$ time steps, we use the average firing rate

\begin{equation}
r = \frac{1}{T}\sum_{t=1}^{T} s_t^{(L)} .
\end{equation}

The resulting rate vector is then mapped to class logits via a linear classifier,
\begin{equation}
z = W_{\mathrm{out}}\, r + b_{\mathrm{out}}.
\end{equation}

In our implementation, the readout head follows the same stabilization pattern as the hidden blocks: we apply batch normalization and dropout to the rate features before the final linear layer. Importantly, the spike-rate readout aggregates the last-layer spike train over the fixed window and produces class logits, which are required both for the cross-entropy training objective and for computing calibrated predictive probabilities (via softmax) and uncertainty metrics.

\subsection{Angularity of the deterministic loss landscape}
\label{subsec:angularity}

The key issue that this work addresses is whether the prediction goal of surrogate gradient SNNs is locally smooth with respect to perturbations of the network parameters. In regular neural networks, changes to the weights of the network cause changes to the output of the network that are also smooth with respect to changes to the loss of the network. However, for SNNs, the forward pass of the network also involves the generation of spikes, and this generation of spikes can change based on the perturbation of the weights of the network.

In order to study this phenomenon, we study slices of the deterministic cross entropy loss function near a trained parameter vector. Let $\hat{\theta}$ be the learned parameter vector, and $d$ be a unit-norm direction in parameter space. We introduce
\begin{equation}
\mathcal{L}(\alpha)
=
\frac{1}{|\mathcal{B}|}\sum_{(x,y)\in\mathcal{B}}
\mathrm{CE}\!\left(y, f_{\hat{\theta}+\alpha d}(x)\right),
\label{eq:det_slice}
\end{equation}
where $\mathcal{B}$ is a fixed set of minibatches and $\alpha$ controls the perturbation magnitude.

The resulting curve has an angular or piecewise irregular form, rather than a smooth form. The loss curve shows abrupt changes in slope and sharp transitions, which is consistent with the underlying model, as the spiking process naturally leads to such effects, even if the surrogate gradients have a smoothing effect on the backpass, as the forward process still depends on the threshold crossings. Neurons can abruptly begin or cease firing, or change the timing of their spikes, and such abrupt changes propagate through the network to the readout.

This gives us direct evidence that the deterministic local loss landscape in SNNs is locally non-smooth and highly sensitive to small parameter perturbations.
\section{Practical Bayesian Learning Method}
\label{sec:ivon_method}
The following sections are tailored to test the hypothesis that the application of Bayesian learning can mitigate the locally angular predictive behavior of deterministic SNNs. The rationale for this hypothesis is based on the observation that the prediction in SNNs with surrogate gradients ultimately depends on the threshold-based event of spikes. Therefore, small changes in the weights can cause the neurons to fire or not to fire, which in turn can propagate over time and cause abrupt changes in the predictive objective. Therefore, the locally angular predictive behavior of deterministic SNNs might remain. In this context, the application of Bayesian learning is of special interest because it replaces the point estimate of the weights with a distribution over the weights and the prediction with the average over the perturbations in the weights. The application of this approach is expected to mitigate the locally angular predictive behavior of SNNs. 

A practical approach to Bayesian deep learning is to approximate the intractable
posterior distribution over the weights with a tractable family of distributions, whose parameters
are optimized during training. A common choice is a mean-field Gaussian approximation, which remains
simple enough for minibatch optimization while still providing uncertainty estimates over the model
parameters. In this work, we adopt this variational perspective and focus on \textit{Improved
Variational Online Newton} (IVON) \cite{shen2024} as a practical method for optimizing such an approximate posterior
in surrogate-gradient SNNs. We first introduce the mean-field formulation and then describe the IVON
updates used in our experiments.

\subsection{Mean-Field Variational Inference}
\label{subsec:mfvi}
Let \(\mathcal{D}=\{(x_i,y_i)\}_{i=1}^N\) denote the training set and \(\theta\) the network
parameters. Since the exact posterior \(p(\theta\mid\mathcal{D})\) is generally intractable,
variational inference introduces a tractable approximation \(q(\theta)\) and optimizes it to be close
to the true posterior. In the mean-field setting, the parameters are assumed independent under
\(q(\theta)\), leading to a diagonal Gaussian approximation
\begin{equation}
q(\theta)=\mathcal{N}(\mu,\mathrm{diag}(\sigma^2)),
\end{equation}
where \(\mu\) and \(\sigma\) are learned variational parameters.

Rather than learning a single point estimate of the weights, variational Bayesian learning seeks a
distribution over plausible parameters by maximizing the evidence lower bound (ELBO),
\begin{equation}
\mathcal{L}_{\mathrm{ELBO}}(\mu,\sigma)
=
\mathbb{E}_{q(\theta)}\!\left[\sum_{i=1}^{N}\log p(y_i\mid x_i,\theta)\right]
-
\mathrm{KL}\!\left(q(\theta)\,\|\,p(\theta)\right),
\end{equation}
or equivalently by minimizing the negative ELBO. The first term encourages good predictive fit to the
data, while the Kullback--Leibler term regularizes the approximate posterior toward the prior
\(p(\theta)\).

The expectation with respect to \(q(\theta)\) is estimated by Monte Carlo sampling. Using the
reparameterization trick, a sample from the approximate posterior can be written as
\begin{equation}
\theta^{(s)}=\mu+\sigma\odot\varepsilon^{(s)},
\qquad
\varepsilon^{(s)}\sim\mathcal{N}(0,I),
\end{equation}
which gives the estimator
\begin{equation}
\mathbb{E}_{q(\theta)}\!\left[\sum_{i=1}^{N}\log p(y_i\mid x_i,\theta)\right]
\approx
\frac{1}{S}\sum_{s=1}^{S}\sum_{i=1}^{N}\log p\!\left(y_i\mid x_i,\theta^{(s)}\right).
\end{equation}
In practice, the sum over the dataset is replaced by a minibatch estimate and the variational
parameters are updated by stochastic optimization.

With an isotropic Gaussian prior \(p(\theta)=\mathcal{N}(0,\lambda^{-1}I)\), the KL term has the
closed form
\begin{equation}
\mathrm{KL}\!\left(q(\theta)\,\|\,p(\theta)\right)
=
\frac{1}{2}\sum_j
\left[
\lambda\left(\mu_j^2+\sigma_j^2\right)
-\log\!\left(\lambda\sigma_j^2\right)-1
\right].
\end{equation}
\begin{figure}[h]
  \centering
  \includegraphics[width=\linewidth]{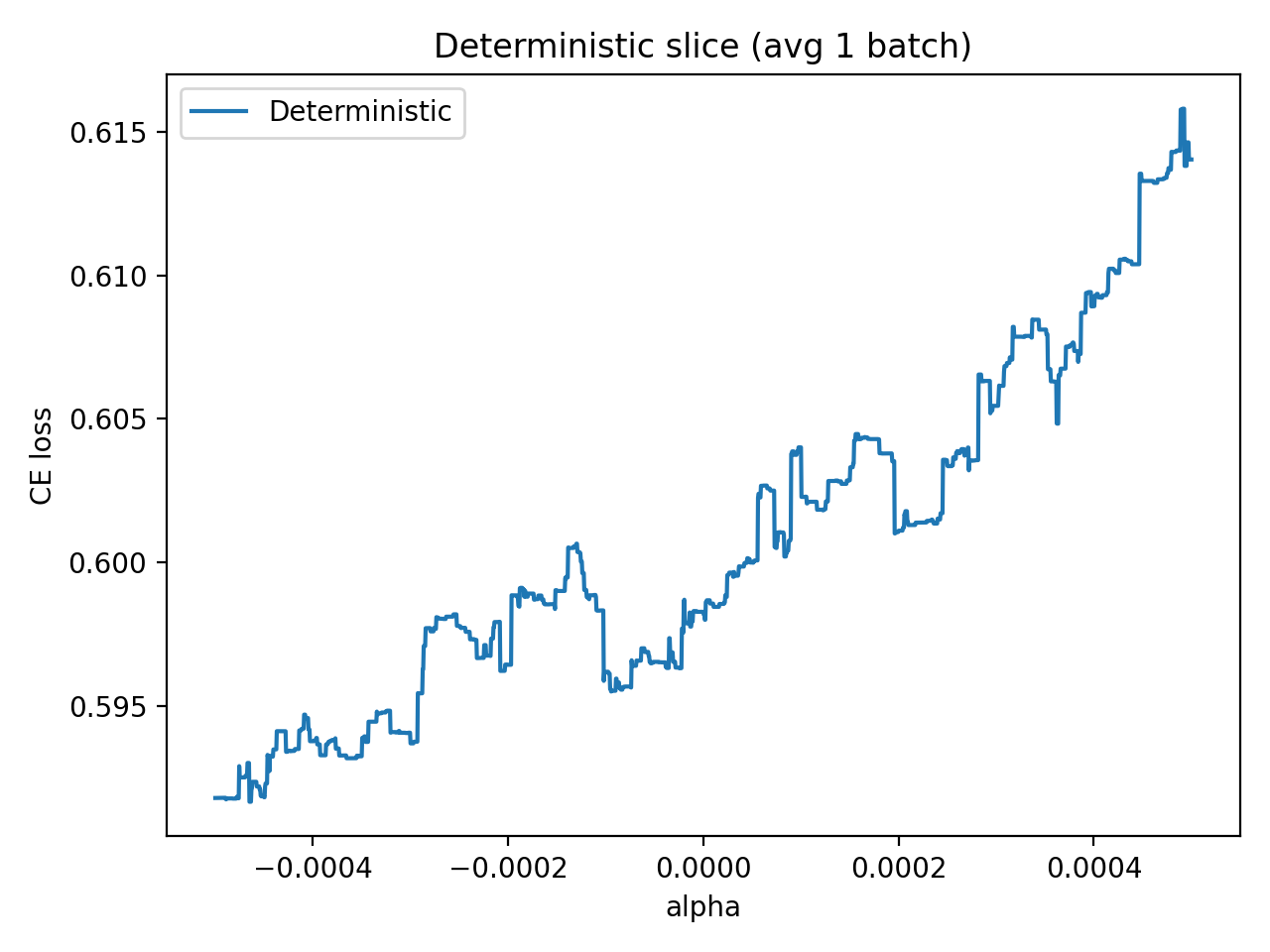}
  \caption{Local one-dimensional slice of the deterministic cross-entropy loss on a fixed batch.}
  \label{fig:det_loss}
\end{figure}

\subsection{IVON: Improved Variational Online Newton}
\label{subsec:ivon}

IVON \cite{shen2024} is a scalable variational method that bridges Bayesian learning and conventional minibatch training. IVON uses a Gaussian approximation for the weights as follows:
\begin{equation}
q_{\mathrm{IVON}}(\theta) = \mathcal{N}(\mu,\Sigma),
\end{equation}
where $\mu$ is used for representing means for the parameters, and $\Sigma$ is for representing weight uncertainty. This is important in our context since we are able to incorporate Bayesian learning into SNN training using surrogate gradient without modifying the SNN structure.

During the training procedure, IVON draws from the current approximate posterior, computes the minibatch gradients, and updates both the posterior mean and the curvature statistic, which regulates the spread of the posterior. Unlike first-order mean-field variational inference, in which the amount of uncertainty in each parameter direction is fixed, IVON uses the curvature statistic to adjust the amount of uncertainty in each parameter direction according to how well that direction is constrained by the data.

IVON is normally used with a Gaussian prior over the weights, which corresponds to $\ell_2$-type regularization with standard scaling. IVON also incorporates an effective sample size, which is represented by $\mathrm{ESS}$. This balances the relative importance of the likelihood and prior depending on the data set size. In our experiments, we used $\mathrm{ESS} = N$, where $N$ is the number of training data.

At test time, predictions are made via Bayesian model averaging. For a given $x^\ast$, $S$ weight configurations are sampled from the learned posterior and the averaged predictive probabilities are used:
\begin{equation}
p(y^\ast\mid x^\ast,\mathcal{D})
\approx
\frac{1}{S}\sum_{s=1}^{S} p(y^\ast\mid x^\ast,\theta^{(s)}),
\qquad
\theta^{(s)}\sim q_{\mathrm{IVON}}(\theta).
\end{equation}

\section{Experimental evaluation of IVON}

All models are trained for 30 epochs. In both deterministic and Bayesian settings, we use a learning-rate scheduler that reduces the step size when validation performance plateaus.

For the deterministic baseline, we use Adam with learning rate $10^{-3}$, batch size 128, and dropout $p_{\mathrm{drop}}=0.1$.

For the Bayesian setting, we keep the same network architecture and replace Adam with IVON. On Heidelberg Digits (HD), IVON uses learning rate $0.1$, batch size 16, dropout $p_{\mathrm{drop}}=0$, and weight decay $10^{-5}$. On Speech Commands (SC), we use the same learning rate and weight decay, with batch size 32 and dropout $p_{\mathrm{drop}}=0$. We additionally freeze the batch-normalization running statistics in the Bayesian model, which we found empirically to be more stable under sampled weight perturbations. We set the effective sample size to $\mathrm{ESS}=N$, where $N$ is the number of training examples. At evaluation time, Bayesian predictions are computed by Monte Carlo averaging on weight configurations $S=20$.

\subsection{Evaluation Metrics}
\label{sec:metrics-results}

We will use the test of predictive quality and uncertainty quality on the validation/test split, using appropriate measures of classification quality, predictive uncertainty, and calibration.

\paragraph{Negative Log-Likelihood (NLL).}
For the predicted probabilities $p_\theta(y \mid x)$, the NLL is given by
\begin{equation}
\mathrm{NLL} = -\frac{1}{n}\sum_{i=1}^n \log p_\theta(y_i \mid x_i).
\end{equation}
Unlike accuracy, NLL checks for both correctness and confidence.

\paragraph{Brier score.}
In addition, we provide the multiclass Brier score, which estimates the squared difference between the predicted probabilities and the corresponding one-hot encoded target probabilities. Let $p_\theta(y=k|x_i)$ be the predicted probabilities for each class $k$ given the input $x_i$. Let $\mathbb{I}[y_i=k]$ be the corresponding one-hot encoded target probabilities. The multiclass Brier score is given by
\begin{equation}
\mathrm{Brier}
=
\frac{1}{n}\sum_{i=1}^{n}\sum_{k=1}^{C}
\left(
p_\theta(y{=}k\mid x_i)-\mathbb{I}[y_i=k]
\right)^2.
\end{equation}
Low values indicate good probabilistic predictions, capturing both accuracy and calibration.

\paragraph{Predictive entropy.}
We use the entropy of the predictive distribution to quantify predictive uncertainty:
\begin{equation}
\mathcal{H}\big[p(y\mid x)\big] = - \sum_{k=1}^{C} p(y{=}k\mid x)\,\log p(y{=}k\mid x).
\end{equation}

\paragraph{Expected entropy and mutual information.}
We use $S$ Monte Carlo samples of the latent variable, denoted by $\{\theta^{(s)}\}_{s=1}^S$ and probabilities $p^{(s)}(y\mid x)=p(y\mid x,\theta^{(s)})$. We compute the expected entropy
\begin{equation}
\mathbb{E}\!\left[\mathcal{H}\big[p(y\mid x,\theta)\big]\right]
\approx \frac{1}{S}\sum_{s=1}^S \mathcal{H}\!\left[p^{(s)}(y\mid x)\right],
\end{equation}
and the predictive mutual information;
\begin{equation}
\mathcal{I}\big(y,\theta \mid x\big)
= \mathcal{H}\!\left[\mathbb{E}_\theta p(y\mid x,\theta)\right]
- \mathbb{E}_\theta\!\left[\mathcal{H}\big[p(y\mid x,\theta)\big]\right],
\end{equation}
where mutual information measures epistemic uncertainty.

\paragraph{Calibration and expected calibration error (ECE).}
We measure the calibration of our model, which means that our confidence should match our actual correctness. We used the expected calibration error (ECE) with confidence bins $B$ to measure the calibration. For each data point, let $\hat{p}_i=\max_k p(y{=}k\mid x_i)$ and $\hat{y}_i=\arg\max_k p(y{=}k\mid x_i)$. Then
\begin{equation}
\mathrm{ECE} = \sum_{b=1}^{B} \frac{| \mathcal{I}_b |}{n}\,
\big|\mathrm{acc}(\mathcal{I}_b) - \mathrm{conf}(\mathcal{I}_b)\big|,
\end{equation}
where $\mathcal{I}_b$ is the set of examples in bin $b$.

\subsection{Results}
\label{sec:results}
Table~\ref{tab:main_results} shows that IVON improves predictive performance over the
deterministic baseline on both datasets. On Heidelberg Digits (HD), where the deterministic model is
already very strong, IVON increases accuracy from \(98.10\%\) to \(99.51\%\) and substantially
reduces both NLL (from \(0.0553\) to \(0.0233\)) and Brier score (from \(0.0294\) to \(0.0105\)).
The entropy is also lower, indicating more concentrated predictions on average, while ECE remains at
a similarly low level. This suggests that, even in a near-saturated regime, IVON improves the
quality of the predictive distribution beyond the gain in accuracy alone.

The effect is more pronounced on Speech Commands (SC), which is the more challenging task. Here,
IVON improves accuracy from \(77.01\%\) to \(79.93\%\), while also reducing NLL, ECE, entropy, and
Brier score. In particular, the drop in ECE from \(0.0224\) to \(0.0133\) indicates better
calibration, and the reduction in Brier score from \(0.3184\) to \(0.2838\) shows that the
predictive probabilities are better aligned with the observed outcomes. The nonzero mutual
information (MI) obtained with IVON on both datasets further indicates the presence of a measurable
epistemic uncertainty component, which is absent in the deterministic point estimate.

Overall, these results suggest that IVON provides benefits beyond raw classification accuracy. It
consistently yields better probabilistic predictions, as reflected by lower NLL and Brier score on
both benchmarks, with the largest overall gains on SC. This is consistent with the view that
learning under weight uncertainty can improve both predictive reliability and uncertainty quality in
surrogate-gradient SNNs.

\begin{table*}[t!]
\centering
\begin{tabular}{llcccccc}
\toprule
Dataset & Method & Acc(\%) $\uparrow$ & NLL $\downarrow$ & ECE $\downarrow$ & Entropy $\downarrow$ & MI $\uparrow$ & Brier $\downarrow$ \\
\midrule
\multirow{2}{*}{HD} 
& Deterministic (MAP) & 98.10 & 0.0553 & 0.0097 & 0.0824 & --     & 0.0294 \\
& IVON (MC $S{=}20$)  & 99.51 & 0.0233 & 0.0103 & 0.0450 & 0.0163 & 0.0105 \\
\midrule
\multirow{2}{*}{SC} 
& Deterministic (MAP) & 77.01 & 0.8063 & 0.0224 & 0.7967 & --     & 0.3184 \\
& IVON (MC $S{=}20$)  & 79.93 & 0.7191 & 0.0133 & 0.6829 & 0.0711 & 0.2838 \\
\bottomrule
\end{tabular}
\caption{Test-set results on Heidelberg Digits (HD) and Speech Commands (SC).}
\label{tab:main_results}
\end{table*}

\subsection{Loss-surface smoothness via 1D weight-space slices}
\label{subsec:smoothness}

The angularity discussion in Section~\ref{subsec:angularity} provides a motivation for investigating whether Bayesian prediction decreases local sensitivity to parameter changes. To this end, we assess one-dimensional slices of the deterministic loss function around a learned solution. Let $\hat{\theta}$ be the learned parameter vector, $d$ be a unit vector in the parameter space, and $\mathcal{B}$ be a set of minibatches. The deterministic slice is
\begin{equation}
\mathcal{L}(\alpha)
=
\frac{1}{|\mathcal{B}|}\sum_{(x,y)\in\mathcal{B}}
\mathrm{CE}\!\left(y, f_{\hat{\theta}+\alpha d}(x)\right),
\label{eq:loss_slice}
\end{equation}
where $\alpha$ controls the perturbation magnitude.

For IVON, we also assess the Bayesian predictive loss calculated by taking the average of the predictive probabilities for the sampled weight configurations according to the learned posterior,
\begin{equation}
\begin{split}
\mathcal{L}_{\mathrm{Bayes}}(\alpha)
&=
\frac{1}{|\mathcal{B}|}\sum_{(x,y)\in\mathcal{B}}
\left[
-\log
\left(
\frac{1}{S}\sum_{s=1}^{S}
p\!\left(y \mid x, \theta^{(s)}(\alpha)\right)
\right)
\right], \\
&\qquad
\theta^{(s)}(\alpha)\sim q(\theta;\alpha).
\end{split}
\label{eq:bayes_slice}
\end{equation}
where $q(\theta;\alpha)$ denotes the IVON posterior centered on the perturbed mean along the slice.

\begin{figure}[h]
  \centering
  \includegraphics[width=\linewidth]{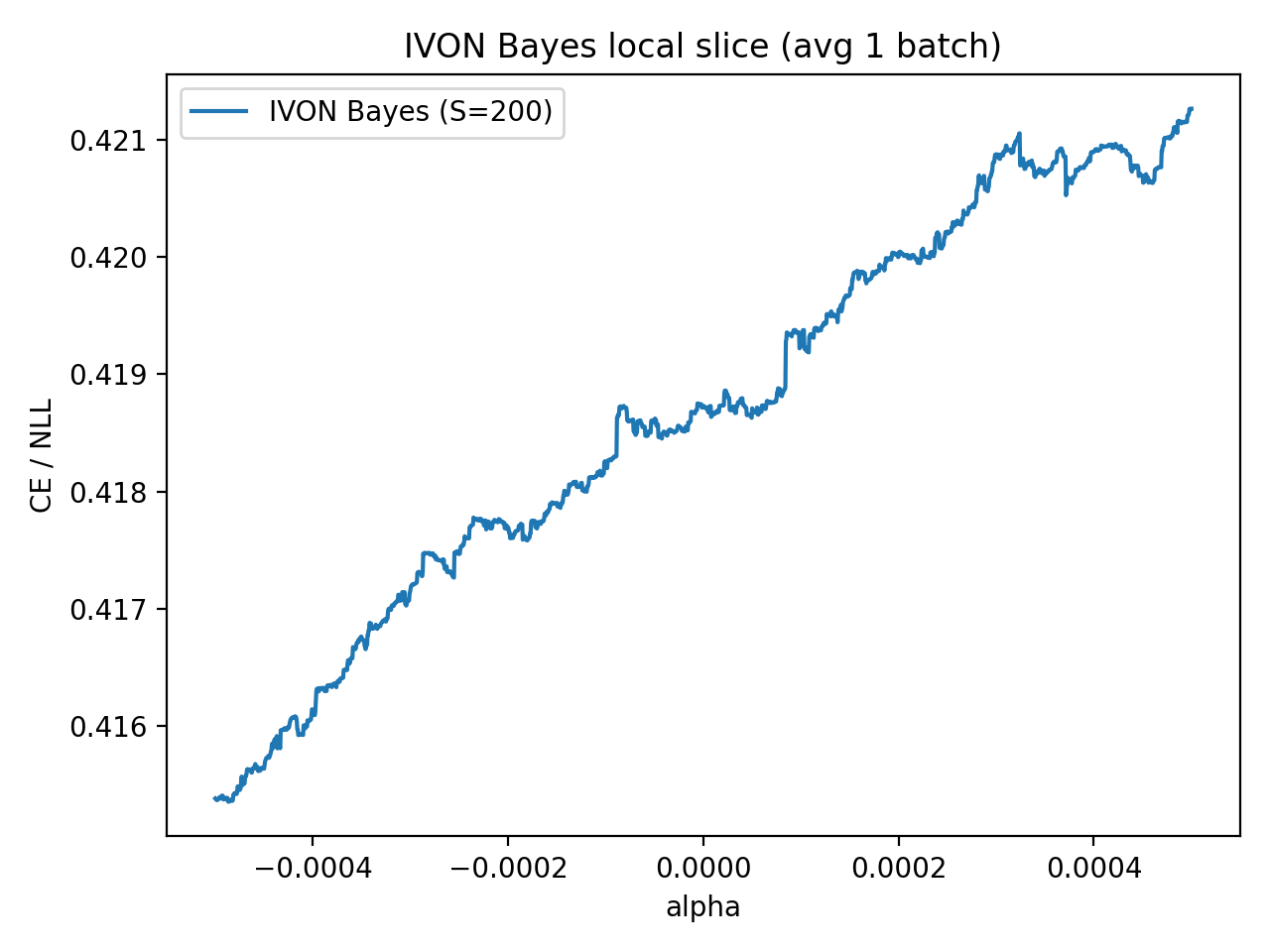}
  \caption{Local one-dimensional slice of the IVON Bayesian predictive loss on a fixed batch.}
  \label{ivon_loss}
\end{figure}

Figure~\ref{ivon_loss} shows the local one-dimensional slice of the IVON Bayesian predictive loss around the learned solution, evaluated on a fixed batch. As the perturbation parameter $\alpha$ varies along the chosen direction, the loss changes in a relatively gradual and regular manner. Although the curve is not perfectly quadratic and still exhibits mild local fluctuations, it is visibly less jagged and less step-like than its deterministic counterpart. This suggests that posterior averaging in IVON attenuates sharp threshold-induced variations in the local predictive objective. In other words, averaging predictions over a distribution of plausible weights softens the effect of abrupt spike-pattern changes, yielding a smoother and less brittle local loss landscape.

Improvements in predictive accuracy, especially in negative log-likelihood, Brier, and uncertainty-based measures, suggest that Bayesian approaches can provide better predictive distributions than deterministic SNN models. In addition, the smoother behavior of the curves in the loss slice analysis is also in line with the idea of accounting for weight uncertainty to alleviate local irregularities due to threshold-based spike generation.

\section{Conclusion}

Combined, the findings indicate that the Bayesian consideration of the uncertainty around the SNN weights has the effect of reducing the local irregularity resulting from thresholded spikes. This is because weight perturbation in deterministic SNNs, due to thresholding, tends to generate locally irregular objectives to predict. On the other hand, Bayesian models take into account the weight uncertainty, which may have the effect of smoothing out the irregularity. In line with this understanding, the Bayesian models performed significantly better at providing good predictive distributions compared to the corresponding deterministic models. In particular, they scored lower in negative log-likelihood, Brier scores, among other measures of uncertainty.
\begin{acks}
We acknowledge the support of the Swiss National Science Foundation under grant number 223717 (\hyperlink{https://data.snf.ch/grants/grant/223717}{MORPhyN}).
\end{acks}

\bibliographystyle{ACM-Reference-Format}
\bibliography{sample-base}

\end{document}